# Data Aware Neural Architecture Search

Emil Njor
emjn@dtu.dk
Technical University of Denmark
Kongens Lyngby, Denmark

Jan Madsen
jama@dtu.dk
Technical University of Denmark
Kongens Lyngby, Denmark

Xenofon Fafoutis
xefa@dtu.dk
Technical University of Denmark
Kongens Lyngby, Denmark

## ABSTRACT

Neural Architecture Search (NAS) is a popular tool for automatically generating Neural Network (NN) architectures. In early NAS works, these tools typically optimized NN architectures for a single metric, such as accuracy. However, in the case of resource constrained Machine Learning, one single metric is not enough to evaluate a NN architecture. For example, a NN model achieving a high accuracy is not useful if it does not fit inside the flash memory of a given system. Therefore, recent works on NAS for resource constrained systems have investigated various approaches to optimize for multiple metrics. In this paper, we propose that, on top of these approaches, it could be beneficial for NAS optimization of resource constrained systems to also consider input data granularity. We name such a system "Data Aware NAS", and we provide experimental evidence of its benefits by comparing it to traditional NAS.

## 1 INTRODUCTION

Manually designing a well-performing Machine Learning (ML) system for a given problem is a challenging task. ML engineers have to rely on knowledge and experience to put together a system that they believe will work well. Afterward, experiments must test the actual performance of the designed system. These experiments can be time-consuming, and therefore the manual design process of Neural Network (NN) systems usually requires a significant amount of engineering work.

An alternative to this is the concept of AutoML. AutoML aims to automate the steps of designing ML systems. AutoML has the potential to both increase productivity and allow non-ML engineers to design ML systems. Neural Architecture Search (NAS) is one of the most well-explored AutoML subfields. In NAS, a computer program automatically searches for an appropriate NN architecture.

It is even more challenging to design well-performing resource constrained ML systems — as in the case of tinyML. tinyML refers to the concept of running ML in ultra-low power systems. Under these power constraints, hardware capabilities are extremely limited. For example, one of the most commonly used tinyML devices — the Arduino Nano 33 BLE Sense is limited to a 64 MHz processor and 256 KB of RAM. Such resource constrained ML systems cannot be optimized for a single metric. For instance, high accuracy for a NN model means nothing if the model does not fit the constraints of the resource constrained device. Therefore, the AutoML methods proposed for resource constrained ML typically try to optimize for more than one metric in various ways.

In this paper, we investigate NAS approaches to resource constrained ML. Specifically, we argue that it is beneficial for NAS approaches for resource constrained systems to also search for an appropriate data granularity. The term "data granularity" refers to the idea that data can be input into an ML model at different levels of granularity. E.g., an audio sample can be input to an ML model at various sample rates. In this situation, we would say that an audio sample given at a sample rate of 24 kHz will be of a higher data granularity than the same audio sample given at 12 kHz. Similarly, an image can be input into an ML model at different resolutions. The number and type of sensors (e.g., mono vs stereo audio) is another example of data granularity.

One of our main insights is that the NN model is not the only part of the entire ML system that consumes the resources of resource constrained devices. In addition, the data given to the model can also use up a significant amount of the already constrained resources. The resource consumption of this data can be regulated by its granularity. Likewise, both the model architecture and the data granularity can have a notable impact on the performance of a ML system. Therefore, both the model and the data present a search space that can be explored to find the best-performing resource constrained ML system for a given problem. It may, e.g., yield a better performing ML system if the data granularity is reduced to leave more resources available for a more complex NN architecture.

Therefore, to investigate this, we propose Data Aware NAS: an expanded NAS for resource constrained ML that considers a unified model and data granularity search space to generate near-optimal ML systems for resource constrained devices.

## 2 DATA AWARE NAS

In this section, we further explore the idea of Data Aware NAS. We first tackle the topic of Single-Objective Optimization (SOO) and Multi-Objective Optimization (MOO) in NAS. We then give our arguments for why including data granularity in the NAS search space can enhance the performance of a system.

### 2.1 Single and Multi Objective Optimization

To use NAS it is necessary to formalize a metric of what a good model is. This is the case in both resource constrained and non-resource constrained NAS settings. This metric will be what the NAS will optimize a NN model for. Traditionally this has been a single metric such as accuracy on a dataset representative of the use case [8]. This is known as SOO NAS. In SOO NAS the goal is simply to return the NN model that optimizes the single metric.

More recent works on NAS have considered how to work with NAS when optimizing for multiple metrics. This is known as MOO NAS [3]. MOO is viewed as more challenging than SOO, as the optimization goal gets distributed on several metrics. This raises questions both for how to compare NN models internally and which models to output as the result of a MOO NAS. One way for MOO NAS to compare models internally to decide which models to keep working on is to create a combined metric from a weighted sum of the metrics [14]. An alternative is to evaluate solutions based on whether they are on the Pareto Frontier [3]. Regardless of the comparison criteria used internally in the MOO NAS, it is common practice to output the Pareto Frontier as the final result. Returning a Pareto Frontier of solutions allows for a use case-specific evaluation on which weighting of the multiple optimization metrics to prefer. MOO is often preferable to SOO in a resource constrained ML setting. This is because compromises between multiple metrics are an inherent part of generating ML models for resource constrained devices.

## 2.2 Benefits of Data Aware NAS

It is paramount to make the most of the available resources when designing resource constrained ML systems. The resources available on a device can refer to various aspects depending on the device. However, there are some resources that we typically care about for resource constrained ML. Some of them are memory consumption, energy consumption and the time it takes to run inference on the model. In this paper, we will focus on memory consumption, as this is often a good proxy for other resources. E.g., a model taking up less memory will (on average) consume less energy and require less time to conduct inference. At the same time, memory consumption is easy to reason about for a variety of devices.

In NN-based resource constrained ML, there are typically two types of data that take up the majority of the memory. One is the NN model, which can be optimized in a MOO NAS. The other one is the data given as input to said model. Consider an example of audio data. Typically, a microphone sensor generates this data at a sample rate. It is then sent to the ML device, which saves it in its memory. At this point, a common step is to preprocess the samples into, e.g., a spectrogram using a Short Time Fourier Transform (STFT), and afterward input it to the ML model. Such a preprocessing step requires a number of samples corresponding to the window of time that one run of the ML model should analyze. Thus, in the best-case scenario, the minimum size of this buffer will at some point be:

$$Size = Sample_{size}(\text{bit}) * Sample\_Rate(\text{Hz}) * Window(\text{s})$$

Let us consider an example scenario with a bit width of 8 bits per sample, a sample rate of 6 kHz, and a time to be captured of 5 seconds. The memory consumed in this scenario would be approximately 30 kB — which can be a considerable amount for a small microcontroller. Furthermore, both the data and the NN model can have a significant impact on the performance of the overall ML system. For example, a small NN model may not be able to generate accurate predictions no matter the data given to it. In the same way, even a large NN model may not be able to generate accurate predictions given data with a very low data granularity. Therefore it is important to balance the resources consumed, and performance gained, by both the data and the model in a resource constrained ML system. This is the goal of our proposed Data Aware NAS. Figure 1 shows an overview of how Data Aware NAS compares to AutoML and NAS.

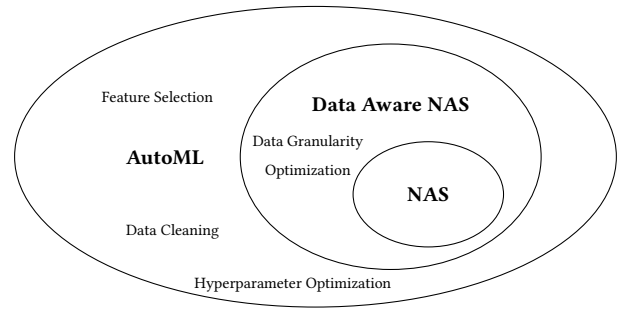

**Figure 1: A visual representation of the relationship between AutoML, Data Aware NAS, and NAS.**

Notably, a paper from 2016 investigates accelerometer datasets and finds that sample rates used in literature were up to 57% higher than needed [12]. It is possible that these results transfer to the field of resource constrained ML and to other types of data and data parameters. If so, this could mean that lowering data granularity could lead to a significant performance gain in resource constrained ML systems.

Lastly, it is worth noting that there exists an interesting interplay between data granularity and NN model size. For typical NNs, the size of the input layer is the same as that of the input data. Therefore, reducing the data granularity will reduce the model size. For some specialized NN models this effect is stronger. For example, for autoencoders, the size of both their input layer and output layer is equal to the size of the input data.

## 3 RELATED WORK

The related work of this paper can be split into three categories:

**Resource Constrained NAS:** Works related to using NAS to find a good NN architecture under some given resource constraints.

**Resource Constrained AutoML:** Works related to creating an automatic process for creating resource constrained ML systems.

**Manual Data Granularity Optimization:** Works where the authors have manually applied optimizations to their data granularity.

## 3.1 Resource Constrained NAS

One of the earliest works on resource constrained NAS was by Elsken et al. [7], who proposed a MOO NAS using evolutionary algorithms. The output of their approach is a Pareto Frontier of possible configurations with respect to multiple parameters e.g. the number of parameters or latency [7]. Around the same time, Hsu et al. proposed a MOO NAS based on reinforcement learning that optimizes both accuracy and the number of MAC operations [11].

A NAS system from Lu et al. does MOO on classification error and the number of floating point operations [15]. A paper by Fedorov et al. uses a combination of NAS and NN pruning to design small but well-performing models [9]. Wang et al. [20] consider an expanded form of NAS for resource constrained ML. Their expanded NAS considers both the neural architecture, quantization, and pruning strategies. To improve search time they use a predictor for model accuracy instead of retraining a model for all combinations. Cassimon et al. [4] propose a MOO NAS using reinforcement learning. In this work, the authors add hard and soft constraints to the reward function of the reinforcement learning model to include multiple objectives. An example of a hard constraint could be that the model should fit within the amount of Random Access Memory (RAM) available. A soft constraint could be the accuracy of the network. A work by Bakhtiarifad et al. describes a NAS system based on a tabular benchmark that is aware of the model's energy consumption [3].

While all of these works propose useful variations of resource constrained NAS, none consider the data granularity optimization proposed in this paper.

### 3.2 Resource Constrained AutoML

Edge Impulse, a well-known company in the tinyML scene has released an AutoML tool known as the EON Tuner. The EON Tuner investigates different combinations of input granularity and NNs architecture [2]. Due to its proprietary nature, we cannot tell how the EON turner proposes its solutions. However, from using and hearing about the software, it is our understanding that it generates some combinations at its initialization stage and evaluates these architectures according to several metrics. As such, the software seems to have no exploitative side where it seeks to improve on initially good architectures.

Doyo et al. recently proposed a paradigm known as tinyML as a Service (tinyMLaaS) [6]. With this, they aim to combat the heterogeneity that comes from multiple ML frameworks having to work together with multiple hardware manufacturers, each with their own tools to deploy tinyML models. Their proposal is to create a platform that contains ML compilers for every combination of framework and hardware available. This platform can then generate tinyML systems for any collection of heterogeneous devices. The platform should furthermore be able to deploy these systems to the heterogeneous devices using Software updates Over The Air (SOTA) technologies. We envision that a Data Aware NAS system, as proposed in this paper, could be of use in such a system.

### 3.3 Manual Data Granularity Optimization

In our prior paper we investigated manually reducing the data granularity for tinyML applications in a predictive maintenance context. This revealed a significant room for reducing the sample rate while preserving model performance and improving inference time [17]. Similarly, a paper by Zalewski et al. [21] describes the manual design of a tinyML system that uses differences in data granularity. This architecture initially uses a low data granularity model for classification. A higher data granularity model then gets activated if the lower data granularity model is not confident in its result [21]. While these papers provide proof of the use of data granularity optimization, they do not propose to automate the concept.

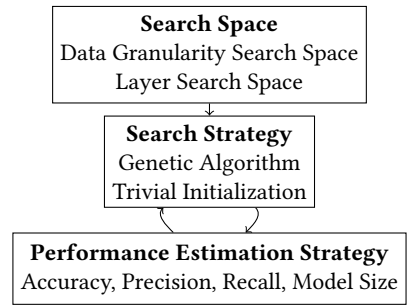

Figure 2: An overview of our system design.

## 4 SYSTEM DESIGN

In this section, we describe the system design of a prototype of a Data Aware NAS. An overarching theme of the design is that we do not attempt to create an optimal Data Aware NAS system but rather try to provide a simple proof of concept system. Therefore, we have often decided in favor of a simple design rather than a complicated one that could have yielded improved performance. Also note that we designed our system to generate Convolutional Neural Networks (CNNs). This is due to CNNs being the target for most NAS research [14]. A quick overview of our system design can be seen in Figure 2.

There are mainly three components in a NAS system: a search space, a search strategy, and a performance estimation strategy [8].

In our system, we split the search space into a search space for data granularity and a search space for the convolutional layers in a CNN. The data granularity search space is heavily dependent on the system use case. Our system design considers this by making the search space configurable. By default, we assume the data to be audio. Since the purpose of the prototype system is only to prove the concept of Data Aware NAS, we have limited the search space to contain values for sample rate and type of preprocessing. The layer search space is less dependent on the use case, but still able to be configured. By default, we choose to include the number of filters in a convolutional layer, the filter size, and the activation function used by that layer.

For the search strategy, we use a genetic algorithm — a common choice for NAS search strategies [8]. Our system supports two types of initializations for the algorithm. By default, we generate an initial population of trivial solutions, which according to [19] has worked well for other NASs based on genetic algorithms. Alternatively, we generate a random initial population. Our search strategy then evaluates the population using our performance estimation strategy described in the following paragraph. Based on tournament selection [14], we chose some of the fittest individuals from this population to be the baselines for the next population. Finally, we apply various crossover and mutation operators to these baselines to generate the next population. Note that these operators need to be implemented according to the configured search space. The default crossover operator of our design chooses two individuals from the baselines and randomly selects values from either to create a new individual. We design several mutation operators for our default search space. See Table 1 for an overview.

Table 1: The default search space parameters and their mutation operators.

| Search Space Parameter | Possible Values | Mutation Operators |
| --- | --- | --- |
| Sample Rate (SR) | 48 000 Hz, 24 000 Hz, 12 000 Hz, 6000 Hz, 3000 Hz, 1500 Hz, 750 Hz, 375 Hz | Increase or decrease one value. |
| Preprocessing Type (PT) | Spectrogram (SP), Mel-Spectrogram (MS), Mel-frequency Cepstrum (MFCC) | Change to another value. |
| Number of Layers (L) | 1,2,3,4,5 | Increase or decrease one value. |
| Number of Filters (F) (per layer) | 2, 4, 8, 16, 32, 64, 128 | Increase or decrease one value. |
| Filter Size (FS) (per layer) | 3, 5 | Increase or decrease one value. |
| Activation Function (AF) (per layer) | Rectified Linear Unit (R), Sigmoid (S) | Change to another value. |

As we are designing a Data Aware NAS for resource constrained ML, we need multiple metrics in our performance estimation strategy. In this proof of concept design, we choose to optimize for Accuracy (Acc), Precision (Pre), Recall (Rec), and Model Size (MS). We choose to use these metrics as accuracy, precision, and recall are fairly easy-to-evaluate metrics that give a good indication of the predictive performance of a system. Model size is another easy-to-evaluate metric for resource constraints that can also act as a reasonable proxy for other resource constraints.

The evaluation of these metrics is not done in the same way. Therefore, our performance estimation strategy consists of two subsystems. The first subsystem evaluates accuracy, precision, and recall metrics by training the given CNN for a configurable number of epochs. The relatively small models used in resource constrained ML mean that we do not suffer from extremely long model training times which has prompted other NAS systems to evaluate these metrics indirectly. The second subsystem evaluates the size of our model by exporting the model to a binary format and recording the number of bytes that the model uses. We use a sum of these metrics to compare two ML systems. To ensure that no one metric dominates the others, we normalize the metrics to a range between 0 and 1 before summing them. Accuracy, precision, and recall natively fall within this range. Model size, on the other hand, is a positive integer with no upper bound. Therefore, we convert model size to a value between 0 and 1 using the formula below:

$$e^{-\frac{Model_{size}}{Approximate\_Model\_Size\_Range}}$$

As the final result, we output the Pareto Frontier of generated ML systems.

## 5 EXPERIMENTS SETUP

This section contains information about the setup used to conduct experiments on the system described in Section 4. We implement our system in the Python 3.10 programming language. See Appendix A for a link to the GitHub repository that hosts the code of this implementation. Python allows us to interface with popular libraries for ML and NNs. Of these libraries, we use the TensorFlow (TF) [5] library to implement CNN architectures, and the TensorFlow Lite (TFL) library to calculate the model size of these architectures. We furthermore use commonly used python libraries such as Numpy [10] for efficient data structures, librosa [16] for audio pre-processing, joblib for easy parallelization, and scikit-learn [18] for evaluation of accuracy, precision, and recall.

Table 2: Values for configurable parameters used in our experiments written in natural language.

| Parameter | Value |
| --- | --- |
| Number of output classes | 2 |
| Loss Function | Categorical Cross Entropy |
| Width of dense layer | 10 |
| Number of normal files to use | 900 |
| Number of anomalous files to use | 200 |
| Frame Size (STFT) | 2048 |
| Hop Length (STFT) | 512 |
| Number of Mel Filter Banks | 80 |
| Number of MFCCs | 13 |
| Maximum number of layers | 5 |
| Optimizer | Adam |
| Number of training epochs per model | 20 |
| Batch Size | 32 |
| Approximate Model Size Range | 100000 |
| Population Size | 10 |
| Ratio of population updated per generation | 0.5 |
| Ratio of crossover to mutations | 0.2 |

For the experiments, we use a subset of the ToyADMOS dataset [13]. The ToyADMOS dataset consists of audio recordings of toys in normal and anomalous operating conditions. The used subset contains recordings of audio channel one of the first case of a toy conveyor belt. An anomalous operating condition is, e.g., that a conveyor belt has a metallic object attached to its belt. The goal of the ML systems produced by our Data Aware NAS is then to learn to classify unseen audio recordings as either normal or anomalous. We additionally mix noise recordings included in the dataset with the normal and anomalous audio recordings to increase the difficulty of the classification.

To run the experiments we use a high-performance computing cluster available at the Technical University of Denmark. The computing nodes that we use contain either a XeonGold6226R or XeonGold6326 server CPU. These nodes also contain a 40 GB and 80 GB Nvidia A100 GPU respectively. See Table 2 for additional information about parameters for other configuration options.

Under this configuration, we run two sets of experiments. We run each set of experiments five times to make our results independent of the effects of a specific random seed. For each run of an experiment, we give a random seed to the random number generator used in our implementation. This random number generator selects, e.g., the type of mutation operator to apply during a population update. We configure each run of the experiments to generate and evaluate 300 combinations of data granularity and model architecture.

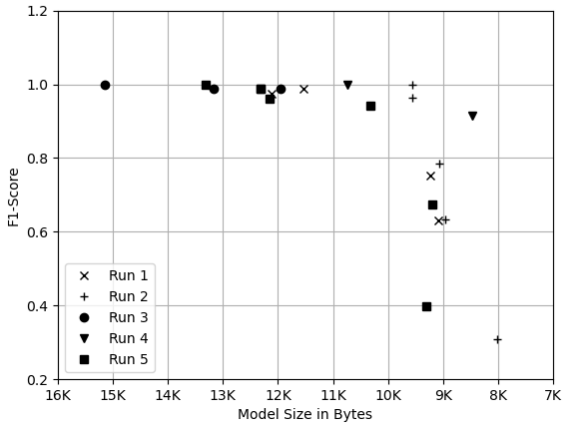

Figure 3: Pareto Frontiers of the Data Aware experiments.

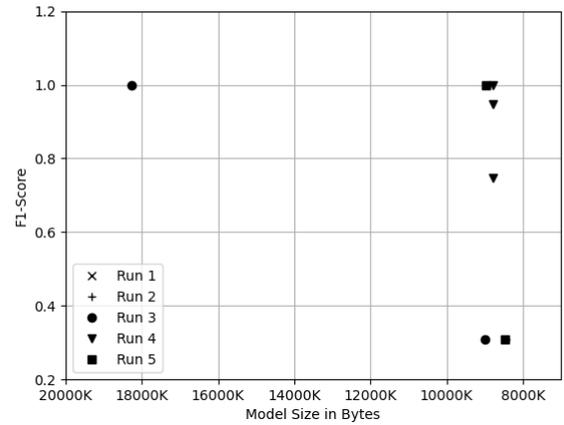

Figure 4: Pareto Frontiers of the Fixed Data experiments.

In the first set of experiments, we run our implementation with the entire search space of all possible values listed in Table 1. We will refer to these experiments as "Data Aware" experiments. The goal of this experiment is to establish a baseline of resource constrained ML system that our Data Aware NAS implementation can generate.

In the second set of experiments, we keep the data granularity at a static configuration. For the static configuration, we choose a sample rate of 6 kHz and a data preprocessing step that generates spectrograms. We will refer to these experiments as "Fixed Data" experiments. By comparing the ML systems generated by this experiment with the systems generated by the Data Aware experiment, we can tell if our Data Aware NAS can generate better ML systems than an equivalent NAS. We believe that a sample rate of 6 kHz and a spectrogram data preprocessing step is likely a configuration that could have been given to a non-Data Aware NAS system. This is due to 6 kHz being a sample rate that balances memory consumption, see Section 2, and data quality [17], and that spectrogram preprocessing is a common preprocessing step for audio data not meant for human ears. Note that due to hardware limitations we could not generate models in this experiment with a number of filters of 128 — since, with this configuration, the model size becomes too large for our processors.

## 6 RESULTS

In this section, we present the results of our experiments. See Table 3 for the Pareto Frontier of ML systems found in our experiments. We also plot the models on the Pareto Frontier of each Data Aware experiment in Figure 3, and each Fixed Data experiment in Figure 4.

From these results, it is clear that both the Data Aware and Fixed Data experiments find accurate ML systems. The ML systems found by our Data Aware experiments, however, have a model size about three orders of magnitudes smaller than the ML systems found by our Fixed Data experiments. As the model size metric does not include the memory consumption of the input data buffer for the ML system, the memory consumption of the two ML systems will in reality be even further apart. This suggests that our hypothesis of being able to create better resource constrained models using a Data Aware NAS than by using a regular NAS is correct.

Surprisingly, the sample rate for the ML systems found in our Data Aware NAS is at either 375 Hz or 750 Hz. These values are low compared to sample rates found to be ideal in a naive manual search [17]. Likewise, it seems that MFCCs are the best type of preprocessing for these ML systems. This is also surprising given that MFCCs were originally designed to help audio processing of human speech — not of industrial machinery. It is therefore also our belief that an engineer would not manually have come up with an as accurate and resource efficient ML system as Data Aware NAS. However, these promising results could partly be due to the search space configured in our default system design. This search space may not allow the regular NAS to produce small models, given our choice of data granularity for the Fixed Data experiments. Additionally, it seems that the ToyADMOS dataset contains much redundant data, given that it is possible to reduce the sample rate to 375 Hz while still preserving perfect accuracy. This makes the dataset a good case scenario for Data Aware NAS, and as such the difference in model size may not be as large for other datasets.

## 7 CONCLUSION AND FUTURE WORK

In this paper, we presented our idea of Data Aware NAS — an expanded form of NAS that additionally searches for an optimal data granularity for the generated NN model. We argue that such a system is especially useful when designing resource constrained ML systems. We design and provide a proof of concept implementation of a Data Aware NAS and conduct experiments to give experimental proof of the validity of our idea. The results confirm our hypothesis that we can create better models under resource constraints when considering data granularity.

An obvious extension of this paper is to expand the current system to include other metrics related to resource consumption.

Table 3: Pareto Frontier of ML systems found in our experiments. We describe each part of the NN architecture according to the component name plus the number of a layer. E.g. FS3 is the filter size of the 3rd convolutional layer. No Pareto Frontier ML system used the possible 5th NN layer.

| Experiment | Run | SR | PT | F1 | FS1 | AF1 | F2 | FS2 | AF2 | F3 | FS3 | AF3 | F4 | FS4 | AF4 | Acc | Pre | Rec | MS |
|---|---|---|---|---|---|---|---|---|---|---|---|---|---|---|---|---|---|---|---|
| Data Aware | 1 | 375 | MFCC | 2 | 3 | R | 2 | 5 | S | - | - | - | - | - | - | 90% | 71% | 80% | 9 240 B |
| Data Aware | 1 | 750 | MFCC | 2 | 5 | S | 2 | 5 | R | 2 | 5 | R | - | - | - | 100% | 98% | 100% | 11 536 B |
| Data Aware | 1 | 375 | MFCC | 2 | 3 | R | 2 | 5 | R | - | - | - | - | - | - | 84% | 55% | 75% | 9 088 B |
| Data Aware | 1 | 750 | MFCC | 2 | 3 | S | 2 | 5 | R | 2 | 5 | R | - | - | - | 100% | 100% | 100% | 16 528 B |
| Data Aware | 1 | 750 | MFCC | 4 | 5 | S | 2 | 5 | R | 2 | 5 | R | - | - | - | 99% | 100% | 95% | 12 108 B |
| Data Aware | 2 | 375 | MFCC | 2 | 5 | R | 2 | 3 | S | - | - | - | - | - | - | 92% | 79% | 78% | 9 064 B |
| Data Aware | 2 | 375 | MFCC | 2 | 5 | R | 2 | 3 | R | - | - | - | - | - | - | 83% | 52% | 83% | 8 960 B |
| Data Aware | 2 | 750 | MFCC | 8 | 5 | R | 2 | 5 | S | 2 | 5 | R | 2 | 3 | R | 100% | 100% | 100% | 9 556 B |
| Data Aware | 2 | 750 | MFCC | 8 | 5 | S | 2 | 5 | R | 2 | 5 | R | 2 | 3 | R | 99% | 93% | 100% | 9 552 B |
| Data Aware | 2 | 750 | MFCC | 4 | 5 | S | 2 | 5 | S | 2 | 5 | R | 2 | 3 | R | 18% | 18% | 100% | 8 012 B |
| Data Aware | 3 | 750 | MFCC | 8 | 5 | R | 2 | 5 | R | 2 | 5 | R | - | - | - | 100% | 100% | 98% | 13 172 B |
| Data Aware | 3 | 750 | MFCC | 8 | 5 | R | 4 | 5 | R | 2 | 5 | R | - | - | - | 100% | 100% | 100% | 15 148 B |
| Data Aware | 3 | 750 | MFCC | 4 | 5 | R | 2 | 5 | R | 2 | 5 | R | - | - | - | 100% | 98% | 100% | 11 956 B |
| Data Aware | 4 | 750 | MFCC | 4 | 5 | R | 2 | 5 | R | 2 | 3 | R | 4 | 5 | S | 100% | 100% | 100% | 10 744 B |
| Data Aware | 4 | 750 | MFCC | 4 | 5 | R | 2 | 5 | S | 2 | 3 | R | 2 | 5 | S | 97% | 90% | 93% | 8 464 B |
| Data Aware | 5 | 375 | MFCC | 2 | 3 | R | 2 | 5 | S | - | - | - | - | - | - | 86% | 60% | 78% | 9 192 B |
| Data Aware | 5 | 375 | MFCC | 2 | 3 | S | 2 | 5 | S | - | - | - | - | - | - | 50% | 26% | 90% | 9 304 B |
| Data Aware | 5 | 750 | MFCC | 8 | 5 | S | 2 | 5 | S | 4 | 5 | R | 2 | 3 | S | 98% | 89% | 100% | 10 324 B |
| Data Aware | 5 | 750 | MFCC | 2 | 5 | R | 4 | 5 | R | 2 | 5 | S | - | - | - | 100% | 98% | 100% | 12 308 B |
| Data Aware | 5 | 750 | MFCC | 2 | 5 | R | 4 | 5 | R | 2 | 5 | S | - | - | - | 100% | 100% | 98% | 12 308 B |
| Data Aware | 5 | 750 | MFCC | 2 | 5 | R | 4 | 5 | R | 2 | 5 | R | - | - | - | 99% | 97% | 95% | 12 156 B |
| Data Aware | 5 | 750 | MFCC | 4 | 5 | R | 4 | 5 | R | 2 | 5 | S | - | - | - | 100% | 100% | 100% | 13 316 B |
| Fixed Data | 1 | 6000 | SP | 64 | 3 | R | 4 | 3 | R | 2 | 5 | S | - | - | - | 100% | 100% | 100% | 8 966 120 B |
| Fixed Data | 1 | 6000 | SP | 64 | 3 | R | 2 | 3 | S | 2 | 5 | S | - | - | - | 81% | 0% | 0% | 8 961 264 B |
| Fixed Data | 2 | 6000 | SP | 16 | 5 | R | 2 | 5 | R | - | - | - | - | - | - | 100% | 100% | 100% | 8 957 760 B |
| Fixed Data | 2 | 6000 | SP | 16 | 5 | R | 2 | 5 | R | 64 | 3 | R | 2 | 5 | R | 18% | 18% | 100% | 8 438 432 B |
| Fixed Data | 2 | 6000 | SP | 16 | 5 | R | 2 | 5 | R | 64 | 3 | R | 2 | 5 | R | 81% | 0% | 0% | 8 438 432 B |
| Fixed Data | 3 | 6000 | SP | 32 | 3 | S | 4 | 5 | S | - | - | - | - | - | - | 100% | 100% | 100% | 18 278 100 B |
| Fixed Data | 3 | 6000 | SP | 32 | 3 | R | 16 | 5 | R | 64 | 3 | R | 2 | 5 | S | 81% | 0% | 0% | 8 697 324 B |
| Fixed Data | 3 | 6000 | SP | 32 | 3 | R | 8 | 5 | S | 2 | 3 | R | - | - | - | 18% | 18% | 100% | 8 981 060 B |
| Fixed Data | 4 | 6000 | SP | 2 | 5 | R | 32 | 5 | R | 2 | 3 | R | - | - | - | 98% | 100% | 90% | 8 782 444 B |
| Fixed Data | 4 | 6000 | SP | 2 | 5 | R | 32 | 5 | R | 2 | 3 | R | - | - | - | 88% | 60% | 100% | 8 782 444 B |
| Fixed Data | 4 | 6000 | SP | 4 | 5 | R | 32 | 5 | S | 2 | 3 | R | - | - | - | 100% | 100% | 100% | 8 789 220 B |
| Fixed Data | 4 | 6000 | SP | 4 | 5 | R | 32 | 5 | S | 2 | 5 | R | - | - | - | 81% | 0% | 0% | 8 613 956 B |
| Fixed Data | 5 | 6000 | SP | 8 | 5 | S | 4 | 3 | S | 64 | 5 | S | 2 | 5 | S | 18% | 18% | 100% | 8 456 952 B |
| Fixed Data | 5 | 6000 | SP | 8 | 5 | S | 4 | 3 | S | 64 | 5 | S | 2 | 5 | S | 81% | 0% | 0% | 8 456 952 B |
| Fixed Data | 5 | 6000 | SP | 32 | 5 | S | 2 | 5 | R | - | - | - | - | - | - | 100% | 100% | 100% | 8 962 744 B |

This could, e.g., be energy consumption, memory consumption, or inference time. We expect that energy consumption could be estimated using a tool such as CarbonTracker [1]. Estimation of memory consumption and inference time could be based on a generated tabular benchmark. However, the tabular benchmark for inference time would have to be generated individually for each device to get accurate estimations.

As described in Section 6, the ToyADMOS dataset might be a good case scenario for Data Aware NAS due to its redundancy. In the future, it could be interesting to investigate Data Aware NAS on other datasets that do not exhibit as much redundancy. Likewise, it could be investigated whether a larger search space for data granularity and NN models could create better ML systems.

Furthermore, it could be interesting to design and implement a more complex but better-performing Data Aware NAS. We could then compare the ML systems generated by this system to ML systems generated using other NAS solutions for resource constrained systems. For example, some systems proposed in Section 3.

It could also be interesting to include this system in a larger AutoML pipeline — e.g., something like the system described in tinyMLaaS [6].

## A  RESOURCES

The implementation of our Data Aware NAS system is available at: https://github.com/Ekhao/DataAwareNeuralArchitectureSearch.

## ACKNOWLEDGMENTS

This work is supported by the Innovation Fund Denmark for the project DIREC (9142-00001B).